\documentclass{article}
\usepackage{spconf,amsmath,graphicx}
\usepackage[hyphens,spaces]{url}
\usepackage{tabularx}
\usepackage{pbox}
\usepackage{threeparttable}

\title{Dementia Assessment Using Mandarin Speech with an Attention-based Speech Recognition Encoder}
\name{  Zih-Jyun Lin$^{1}$,
        Yi-Ju Chen$^{1}$, 
        Po-Chih Kuo$^{1}$, 
        Likai Huang$^{2}$, 
        Chaur-Jong Hu$^{2}$, 
        Cheng-Yu Chen$^{3,4}$}
\address{$^{1}$Department of Computer Science, 
        National Tsing Hua University, 
        Hsinchu, Taiwan\\
        $^{2}$Dementia Center and Department of Neurology, Shuang-Ho Hospital, New Taipei City, Taiwan\\
        $^{3}$Research Center for Artificial Intelligence in Medicine, 
        Taipei Medical University, Taipei, Taiwan\\
        $^{4}$Department of Medical Imaging, Taipei Medical University Hospital, Taipei, Taiwan}
\begin{document}
%
\maketitle
\begin{abstract}
Dementia diagnosis requires a series of different testing methods, which is complex and time-consuming. Early detection of dementia is crucial as it can prevent further deterioration of the condition. This paper utilizes a speech recognition model to construct a dementia assessment system tailored for Mandarin speakers during the picture description task. By training an attention-based speech recognition model on voice data closely resembling real-world scenarios, we have significantly enhanced the model's recognition capabilities. Subsequently, we extracted the encoder from the speech recognition model and added a linear layer for dementia assessment. We collected Mandarin speech data from 99 subjects and acquired their clinical assessments from a local hospital. We achieved an accuracy of 92.04\% in Alzheimer’s disease detection and a mean absolute error of 9\% in clinical dementia rating score prediction. \footnote{Code Available at https://github.com/jason7580/End-to-End-ASR-and-Dementia-detection-system}
\end{abstract}
\section{Introduction}
There are over 55 million people worldwide with dementia, and it is projected to grow to 139 million by 2050 \cite{who}. Dementia is a complex condition that cannot be diagnosed through a single test. Instead, clinicians utilize a variety of assessments, including psychiatric evaluations, cognition, and brain imaging, to identify cognitive impairment associated with dementia. The mini-mental state examination (MMSE) and the clinical dementia rating (CDR) are two commonly employed cognitive assessment tools. Furthermore, a variant of the CDR, known as the Clinical dementia rating sum of boxes (CDR-SOB), is more quantitative and involves summing up scores from different categories, providing a more detailed assessment of dementia. 


Language assessment tools have been recognized as low-cost and effective instruments with high specificity and sensitivity in diagnosing dementia at its earliest stages. They excel in identifying language impairments, which are among the first cognitive signs of various forms of dementia. One type of language assessment tool is the picture description task. The picture description task can examine the subject's structural language skills and assess the subject's general cognition and perception. The Cookie Theft (CT) picture description test involves asking subjects to describe a picture, recording their responses, and assessing them based on three factors: Cookie, Fluency, and Recall, with the objective of evaluating whether they suffer from Alzheimer's disease (AD) or mild cognitive impairment (MCI) \cite{Cummings}.

As various speech recognition systems become more mature, the ability to recognize speech is improving. However, speech recognition remains challenging for elderly adults or individuals with speech impairments. The success of speech recognition makes automatic language assessment possible. Most dementia-related speech recognition research still primarily uses English as the main research language \cite{relateZ}. In addition, a common classification method involves obtaining speech features by combining both low-level acoustic features extracted from speech signals and high-level linguistic features extracted from speech transcripts \cite{relateY}.

In recent years, research on Chinese-language studies has gradually developed. Some studies have used English training to assess Chinese-speaking patients \cite{relateA}, while others have employed not only speech data \cite{relateD, relateJ} or used feature selection \cite{relateB} for classification. Some studies have utilized the popular approach of pretraining for classification, such as using raw speech as input data, retaining the lower layers of an automatic speech recognition (ASR) model, and adding a simple fully connected neural network on top for dementia classification \cite{relateC,relateG}. However, few studies have used Taiwan's Mandarin as input data so far.

This study aims to develop a dementia assessment system using a Taiwan Mandarin speech recognition model (Figure~\ref{fig:overview}). The model can be used for AD detection and dementia-related rating prediction. We also compared the results with the representation learning framework, the HuBERT model \cite{hubert}.
\label{sec:intro}

\section{Datasets}
\label{sec:datasets}

\subsection{Local Dataset}
\label{ssec:subhead21}
The participants were recruited from the Dementia Center and Department of Neurology, Shuang-Ho Hospital. The Inclusion criteria are: (1)Patients diagnosed as suspected dementia or confirmed with mild dementia, with a CDR score less than 1. (2)Individuals seeking medical attention due to acute stroke, Parkinson's disease, or other neurological disorders. (3)Non-patients who are willing to participate (such as peer relatives or caregivers). Participants themselves or their legal representatives who voluntarily decided to participate in this trial are able to cooperate in signing the written informed consent. Individuals who were unable to cooperate in the study due to severe dementia or other illnesses were excluded from this study. Participants underwent picture description assessments, paragraph reading oral tests, and gait analysis. 

A total of 99 individuals have participated in this study, ranging in age from 23 to 96 years old. Among them, 40 individuals were diagnosed with dementia and 59 individuals without dementia. The dementia group includes patients diagnosed with AD, Dementia with Lewy Bodies, and Frontotemporal Dementia within the past two years. This dataset also includes various scores and features of the participants, including age, education level, cognitive abilities screening instrument (CASI), MMSE, CDR, and CDR-SOB. Our study only utilized CT picture description task segments for training and MMSE, CDR, CDR-SOB, and dementia diagnosis for prediction.
\begin{figure}[htb]

\begin{minipage}[b]{1.0\linewidth}
  \centering
  \centerline{\includegraphics[width=7.5cm]{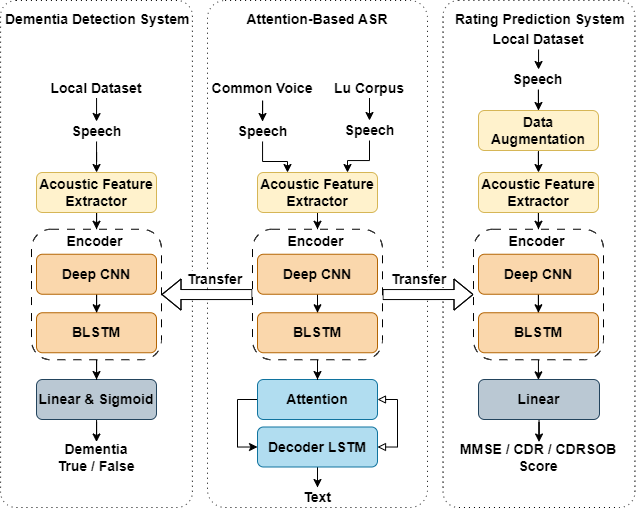}}
\end{minipage}
\caption{System overview}
\label{fig:overview}
\end{figure} 

We excluded 10 participants who used Taiwanese for more than 50\% of the test, as well as one participant with background conversation noise exceeding their voice volume. Therefore, our final classification dataset consists of 88 CT picture description test segments, including 30 segments from dementia patients and 58 segments from non-dementia individuals. The detailed demographic information and relative ratings are shown in Table~\ref{table:dem}.

\begin{table*}[ht]

\centering 
\caption{Demographic information of participants}
\begin{tabular}{|l|l|l|l|l|l|}
\hline
               &        & Overall     & Dementia    & Non-dementia  & p-value          \\ \hline
n              &        & 88          & 30          & 58            &                  \\ \hline
Age, mean (SD) &        & 63.9 (16.4) & 75.0 (8.1)  & 58.2 (16.8)   & \textless{}0.001 \\ \hline
Gender, n (\%) & Female & 58 (65.9)   & 16 (53.3)   & 42 (72.4)     & 0.120            \\ \hline
               & Male   & 30 (34.1)   & 14 (46.7)   & 16 (27.6)     &                  \\ \hline
MMSE, mean (SD) &        & 26.7 (6.0)  & 20.3 (6.7)	& 30.0 (0.3)	& \textless{}0.001 \\ \hline
CDR, mean (SD) &        & 0.2 (0.3)	  & 0.6 (0.3)	& 0.0 (0.0)     & \textless{}0.001 \\ \hline
CDR-SOB, mean (SD) &    & 0.8 (1.6)   & 2.2 (2.1)	& 0.0 (0.0)     & \textless{}0.001 \\ \hline
\end{tabular}
\label{table:dem}
\end{table*}

\subsection{Lu Corpus}
\label{ssec:subhead22}
The Lu corpus \cite{AphasiaBank} contains data from 51 participants, including CT picture description task, all with a Taiwanese Mandarin accent. Based on our inference, the participants likely have varying degrees of dementia or aphasia. We segmented these transcripts into 903 utterances. These segments were used for training the speech recognition system.
\subsection{Common Voice Corpus 11.0} 
\label{ssec:subhead23}
Common Voice \cite{commonvoice} is a publicly available voice dataset contributed by volunteers from around the world. We selected the Mandarin (Taiwan) Common Voice Corpus version 11.0, which consists of a total recording time of 113 hours. We derived our own train, dev, and test subsets, consisting of 89,455, 14,709, and 14,709 utterances, from their train, dev, test, and other subsets.


\section{methods}
\label{sec:pagestyle}

Figure~\ref{fig:overview} shows the proposed method's overview. We used Common Voice and Lu Corpus to train the ASR model and transfer the encoder to train the classification and prediction model with local data. 

\subsection{Attention-based end-to-end ASR}
\label{ssec:subhead31}
This work is based on the established approach of end-to-end sequence-to-sequence (seq-to-seq) ASR \cite{7472621}. To begin, the frame-level acoustic features of each input utterance, including filter bank outputs, deltas, and delta-deltas, were extracted using a sequence encoder. These extracted features were then decoded by a sequence decoder, resulting in a word-level latent representation.

For the encoder, a VGG network \cite{vgg} and a 4-layer bidirectional long short-term memory (LSTM) with a hidden size of 1024 in each direction were utilized. The decoder consisted of a double-layer LSTM with a size of 1024 and employed location-aware attention \cite{locatt}. The word transform layer involved a single-layer linear transformation followed by a Softmax activation. For the tokenizer, Chinese characters were used as the sub-word units.

We selected the Common Voice and Lu corpus as the training data, specifically choosing data with Taiwanese Mandarin accents to make the model more familiar with Taiwanese Mandarin. We utilized the Common Voice dataset because it consists of recordings contributed by individuals and includes background noise that is more typical of our data. Additionally, we incorporated the Lu corpus for training because it contains CT description tasks that are similar to our data, including relevant vocabulary, and primarily involves older adults, individuals with dementia, and individuals with speech impairments. This helps enhance the recognition performance of our model on our data.

\subsection{ASR-based dementia detection system}
\label{ssec:subhead32}
We extracted the encoder from the trained ASR model in the detection system. We added a linear layer on top of the encoder, followed by a Sigmoid activation function, to obtain the classification results (see Figure~\ref{fig:overview}). The input data consisted of filter banks, deltas, and delta-deltas from the CT description task, with each segment associated with a single participant.

We employed two training approaches for our classification system. The first approach involves freezing all layers except for the last layer of the bidirectional LSTM and the final linear layer during training (-H). The second approach involves freezing all layers except for the first convolutional layer, the last layer of the bidirectional LSTM, and the final linear layer during training (-H-L). Both methods were trained using the two different ASR systems mentioned earlier. Finally, we used a threshold of 0.5 on the Sigmoid output for classification.


\subsection{HuBERT-based dementia detection system}
\label{ssec:subhead33}
In this detection system, we utilized Tencent GameMate's speech pre-trained Chinese-HuBERT-Large \footnote{\url{https://huggingface.co/TencentGameMate/chinese-hubert-large?doi=true}} as a base. Similar to our transfer-learning-based dementia detection system, we connected the HuBERT \cite{hubert} output to a linear layer and a Sigmoid activation function to obtain the classification results. The input data consisted of raw audio from the CT description task, provided on a per-participant basis. We only trained the final linear layer. Finally, we used a threshold of 0.5 on the Sigmoid output, where values greater than the threshold were classified as dementia, and values lower than the threshold were classified as non-dementia patients.


\subsection{ASR-based rating prediction system}
\label{ssec:subhead34}
Similar to the classifier architecture, we extracted the encoder from the trained ASR model in the detection system and added a linear layer on top of the encoder. The difference is that we removed the Sigmoid activation function from the final layer to obtain the regression score (see Figure~\ref{fig:overview}). The input data consisted of raw audio from the CT description task, provided on a per-participant basis. Additionally, we addressed data imbalance through data augmentation techniques such as noise-addition, time-shifting, time-stretching, and pitch-shifting. Finally, we froze all layers except for the first convolutional layer, the last layer of the bidirectional LSTM, and the final linear layer during training for our score prediction system.

\subsection{Implementation}
\label{ssec:subhead35}
We used 16-bit, 16,000Hz, mono WAV files as input for all of our models. We use AdaDelta as the optimizer. All models were trained using an RTX 3090 GPU. The ASR model was trained for approximately 140-145 epochs. The ASR-based dementia detection system was trained for 150-175 epochs, while the non-transfer learning version took 19 epochs. The HuBERT-based dementia detection system attained its best results at 88th epoch.

\section{Results}
\label{sec:result}
\subsection{Attention-based ASR}
\label{ssec:subhead41}

We manually transcribed the speech from four participants for validation. In both the training models using only the Common Voice dataset and the one incorporating the Lu corpus, the character error rate (CER) for our participants decreased from 253.35\% to 85.21\%. 

    

\subsection{Dementia detection system}
\label{ssec:subhead42}
Due to the limited amount of data, we conducted a 5-fold cross-validation approach to validate our classification system. The data were evenly divided into five groups, with dementia and non-dementia patients separated, for cross-validation. The experiment included the ASR-based dementia detection system, the dementia detection system trained from scratch, the HuBERT-based dementia detection system, and the SVM-based dementia detection system trained using eGeMAPS \cite{eGeMAPS} and ComParE-2016 \cite{ComParE_2016} acoustic features with linear kernel. We compared the results of the latter three with our ASR-based model. For the ASR-based system, experiments were conducted using ASR models trained with and without the Lu corpus (w/ Lu and w/o Lu). Additionally, experiments were conducted using variations of the ASR model architecture, specifically unfreezing the first layer of the encoder (-H) and unfreezing both the first and last layers of the encoder (-H-L). The results, including sensitivity (SEN), specificity (SPE), area under the curve of ROC (AUC), and confidence intervals (CI) of each metric, are presented in Table~\ref{table:DD}.


\begin{table}[ht] 
\begin{threeparttable}
\footnotesize
\centering
\caption{Performance of Alzheimer’s disease detection}
\begin{tabular}{lllll}
\hline
                        &ACC    &SEN    &SPE    &AUC  \\ 
\hline
W/ transfer             &&&&\\
\hspace{3mm}-L w/o Lu   &0.89   &0.78   &0.95   &0.93 \\
                        &[0.83-0.93]    &[0.66-0.89]    &[0.89-0.98]    &[0.87-0.97]\\
\hspace{3mm}-H-L w/o Lu &0.89   &0.78   &0.95   &0.93 \\
                        &[0.83-0.94]    &[0.66-0.89]    &[0.89-0.98]    &[0.87-0.98]\\
\hspace{3mm}-L w/ Lu    &0.91   &0.87   &0.93   &0.97 \\
                        &[0.85-0.96]    &[0.76-0.96]    &[0.87-0.98]    &[0.95-0.99]\\
\hspace{3mm}-H-L w/ Lu  &0.92   &0.87   &0.95   &0.98 \\
                        &[0.88-0.97]    &[0.77-0.97]    &[0.89-98]    &[0.96-0.99]\\
W/o transfer            &0.89   &0.77   &0.95   &0.97 \\
                        &[0.83-0.94]    &[0.63-0.88]    &[0.89-0.98]    &[0.94-0.99]\\
HuBERT                  &0.84   &0.63   &0.96   &0.90 \\
                        &[0.77-0.90]    &[0.50-0.77]    &[0.92-1.0]    &[0.83-0.96]\\
SVM               &&&&\\
\hspace{3mm} eGeMAPS          &0.72   &0.59   &0.79   &0.69 \\
                        &[0.63-0.78]    &[0.42-0.71]    &[0.69-0.88]    &[0.58-0.76]\\   
\hspace{3mm} ComParE     &0.77   &0.63   &0.86   &0.74 \\
                        &[0.69-0.85]    &[0.48-0.78]    &[0.77-0.93]    &[0.66-0.83]\\

\hline
\end{tabular}
\begin{tablenotes}
\item[1] -H: Unfreeze the first layer of the encoder
\item[2] -L: Unfreeze the last layer of the encoder
\end{tablenotes}
\label{table:DD}
\end{threeparttable}

\end{table}
\subsection{Rating prediction system}
\label{ssec:subhead43}
\begin{figure}[htb]
\begin{minipage}[b]{1.0\linewidth}
\centering
\centerline{\includegraphics[width=8.5cm]{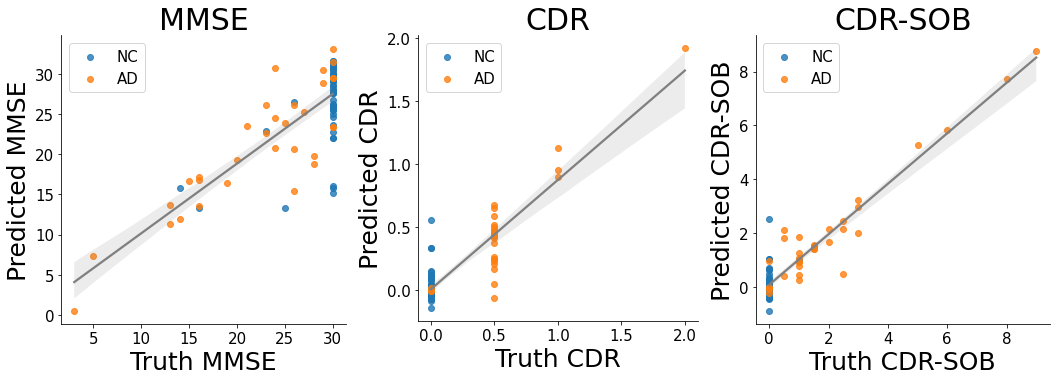}}
\end{minipage}
\caption{Results of rating prediction system}
\label{fig:reg}
\end{figure}

Similar to the validation method for the classification system, we conducted a 5-fold cross-validation approach to validate our score prediction system. For the ASR-based system, experiments were conducted using ASR models trained with the Lu corpus. The final results and their metrics, such as mean absolute error (MAE), mean square error (MSE), root mean square error (RMSE), R squared (R$^2$), and explained variance score (EVS), are presented in Table~\ref{table:RP}, where it could be observed that the regression performance was similar, and the CDR regression resulted outperforming the other two (see Figure~\ref{fig:reg}).

\begin{table}[ht]
\small
\centering
\caption{Performance of dementia-related rating prediction}
\begin{tabular}{llllll}
\hline
            &MAE    &MSE    &RMSE   &R$^2$  &EVS\\
\hline
MMSE        &3.02   &20.5   &4.48   &0.45   &0.56\\
CDR         &0.09   &0.02   &0.15   &0.81   &0.81\\
CDR-SOB     &0.32   &0.29   &0.54   &0.89   &0.89\\
\hline
\end{tabular}
\label{table:RP}
\end{table}

\section{Discussion}
\label{sec:discuss}

We have observed an accuracy gap between the utilization of Hubert and ASR transfer learning. This distinction can be attributed not solely to potential variations in model architecture but also to the utilization of models trained on distinct accents for transfer learning. Furthermore, our transfer learning experiments, whether with or without the incorporation of the Lu corpus for ASR training, demonstrate that ASR models trained with pertinent vocabulary significantly influence the classification outcomes. Finally, the conventional feature extraction-based classification methods produced subpar results when compared to the proposed model. We posit that this discrepancy arises from the fact that the proposed model effectively integrates temporal information. By preserving this temporal information, the model can extract features pertaining to speech patterns, textual content, and intonation.

The experiments show that when we unfreeze both the first and last layers of the encoder in our models, we get better accuracy. This suggests that unfreezing the first layer helps the model adapt to different accents and contexts in the data, improving accuracy. Additionally, it's worth mentioning that using the same model without transfer learning also gave good results, with an accuracy of 88.6\%. This shows that the model can be trained effectively using patient diagnostic recordings and can be useful for monitoring dementia and generating cognitive assessment scores.




\section{Conclusion}
\label{sec:conclude}
This paper presents a method for training an ASR system tailored to older adults and dementia patients. Our approach enhances the recognition of CT picture description tasks and further predict cognitive assessment scores. Furthermore, we apply a relatively simple technique to develop an efficient dementia identification system that allows rapid prediction of classification outcomes and cognitive function scores.

\vfill\pagebreak

\bibliographystyle{IEEEbib}
\bibliography{strings,refs}

\end{document}